\pdfoutput=1
\documentclass[11pt]{article}

\usepackage[final]{EMNLP2022}

\usepackage{times}
\usepackage{latexsym}

\usepackage{graphicx}

\usepackage[T1]{fontenc}
\usepackage[utf8]{inputenc}

\usepackage{microtype}

\usepackage{inconsolata}

\usepackage{amsmath}
\usepackage{amsfonts}

\title{Prompt Compression and Contrastive Conditioning for\\ Controllability and Toxicity Reduction in Language Models}

\author{David Wingate \\
  Brigham Young University\thanks{\hspace{0.05in} Work done while at Nvidia, Inc.}\\
  \texttt{wingated@cs.byu.edu} \\\And
  Mohammad Shoeybi \\
  Nvidia, Inc. \\
  \texttt{mshoeybi@nvidia.com} \\\And
  Taylor Sorensen \\
  University of Washington\thanks{ \hspace{0.05in} Work done while at Brigham Young University} \\
  \texttt{tsor13@cs.washington.edu} \\
  }

\begin{document}
\maketitle

% ===============================================================
% ===============================================================
% ===============================================================

\begin{abstract}
We explore the idea of compressing the prompts used to condition language models, and show that compressed prompts can retain a substantive amount of information about the original prompt. For severely compressed prompts, while fine-grained information is lost, abstract information and general sentiments can be retained with surprisingly few parameters, which can be useful in the context of decode-time algorithms for controllability and toxicity reduction. We explore contrastive conditioning to steer language model generation towards desirable text and away from undesirable text, and find that some complex prompts can be effectively compressed into a single token to guide generation. We also show that compressed prompts are largely compositional, and can be constructed such that they can be used to control independent aspects of generated text. 
\end{abstract}

% ===============================================================
% ===============================================================
% ===============================================================

\section{Introduction}

Language models (LMs), such as GPT-2 \cite{radford2018improving,radford2019language}, BERT \cite{devlin2018bert}, T5 \cite{2020t5}, or GPT-3  \cite{brown2020language}, exhibit a remarkable ability to capture patterns of grammar, vocabulary, cultural knowledge, and conversational rhythms present in natural language. Formally, a LM is a conditional distribution over tokens $p(x_t|x_1,\cdots,x_{t-1})$, with each token $x_t \in \mathcal{V}$ for some vocabulary $\mathcal{V}$.  Throughout this paper, we will refer to $x_h=x_1,\cdots,x_{t-1}$ as the \emph{prompt}.

\begin{figure}
\includegraphics[width=1\linewidth]{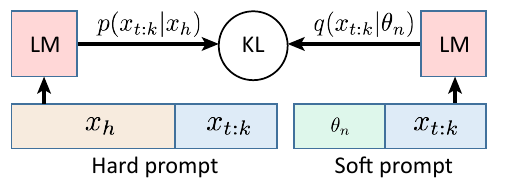}
\caption{Schematic of prompt compression. Weights of the soft prompt are tuned to minimize the KL divergence between hard and soft prompts, for all $x_{t:k}$.}
\label{fig:prompt_compression}
\end{figure}

This paper explores \emph{prompt compression}: the idea that the text $x_h$ used to condition a LM can be approximately represented by a much smaller set of carefully chosen weights, using the framework of soft prompts \cite{lestersoftprompt}. We begin by establishing some basic properties of compressed prompts, and importantly show that while highly compressed prompts lose fine-grained information about the prompt, they can retain general, abstract information.  This motivates our central application: to use such compressed prompts in a Bayesian attribute framework to steer text generation, with specific application to toxicity reduction.

To motivate this more deeply, we briefly sketch how compressed prompts can be used in toxicity reduction. Efforts to reduce toxicity and bias generally follow one of two strategies: the first is to train or fine-tune LMs on carefully curated data, either tagging or labelling it in special ways \cite{salesForceCTRL,lu2022quark} or using data known to be ``clean''. The second is to "steer" the generation of token probabilities away from toxic generations \cite{krause2020gedi,liu2021dexperts}, and towards text with known, desirable properties.

Following previous work, we steer LM probabilities by using a Bayesian attribute classifier framework that involves scoring candidate tokens with different experts.
As an independent contribution, we explore the idea of simply using conditioning text to construct such experts by leveraging the few-shot modeling abilities of LMs \cite{radford2019language,brown2020language}: given a few examples of text containing a pattern of interest, language models are capable of ``analyzing'' such examples and assign high probability to subsequent text exhibiting the same pattern.  Thus, in the same way that language model can, for example, classify the sentiment of a tweet, we use LMs to analyze the toxicity of candidate generations in real-time.  Our method can be considered an exemplar-based method of defining experts that capture desirable and undesirable attributes of generated text.
We term this technique \emph{contrastive contexts}, and note that it reduces the problem of creating experts to one of prompt engineering \cite{reynolds2021prompt}.

%The intuition behind our method is simple: we evaluate candidate text to be generated in two different contexts, one of which is toxic, and one which is benign. If the candidate text is more likely in the toxic context---in other words, if the language model deems it ``similar'' in tone or content to toxic content---then we reduce the probability of that particular generation.  

%In contrast to other methods of detoxification, our method does not require clean datasets, complex generation algorithms, access to model weights, or an ability to backprop through a model.  It is also computationally efficient, and typically involves one additional forward pass through the LM.

However, our conditioning contexts are quite large, which motivated this work. We use prompt compression to mimic an uncompressed prompt (hereafter referred to as "hard" prompt) as closely as possible, thereby saving both computation and space in the context window. Our results demonstrate that this can be very effective, and, in a very surprising finding, that complex prompts can be reduced to a single token and still be useful for toxicity reduction, often with better fluency compared to hard prompts.

The contributions of this paper are three-fold: first, we introduce and formalize the idea of prompt compression; second, we introduce and formalize the method of contrastive contexts in the Bayesian attribute framework; third, we experimentally evaluate our methods, and refine the technique based on various empirical observations, and contribute a careful study of effectiveness as model size varies.

% ===============================================================
% ===============================================================
% ===============================================================

\section{Background and Related Work}
\label{sec:background}

To the best of our knowledge, this is the first work to directly explore prompt compression. However, our work is based on the original soft prompt ideas of \citep{lestersoftprompt}. It is also somewhat related to distillation, where one model is trained to mimic another by matching logits~\cite{distillation_2021}.

% TODO - one of the reviewer's wanted us to add more context into soft prompts, as they say the work is not "self-contained." Is this the best place to put it? Should we include it at all?
Usually, LMs take text as input, which is then tokenized into discrete tokens by a tokenizer. Each token is then mapped to a learned embedding, which is used as input to a transformer \cite{vaswani2017attention}. The idea of soft prompts \cite{lestersoftprompt} is to bypass the need to use discrete tokens with pre-trained embeddings and instead directly learn a series of embeddings via backpropagation. These learned embeddings are then fed directly to the transformer and do not need to correspond to any actual language tokens.

As the centerpiece application of prompt compression, we explore generative controllability \cite{CTRL} and toxicity reduction in language models. 

Our method is most closely related to decode-time algorithms, such as GEDI \cite{krause2020gedi}, which uses Bayes' rule and discriminative models to steer the generation towards a certain attribute; and PPLM \cite{pplm2019}, which uses an estimated gradient with respect to the desired attribute to steer the LM's internal representation at generation time.

Other methods are based on fine-tuning language models with the classical language modeling objective to steer generation. DEXPERTs \cite{liu2021dexperts} combines experts and anti-experts in a product of experts model to reduce toxicity of LMs.

Additionally, reinforcement learning approaches show strong performance at steering language models \cite{learningtosummarize}. By providing rewards, methods such as PPO \cite{schulman2017proximal} and Quark \cite{lu2022quark} represent the current best performance at reducing LM toxicity while maintaining fluency. These methods, however, require a predetermined reward function, which may or may not be feasible depending on the context.

% Decode-time algorithms
% \cite{NPI2020}
% GEDI \cite{krause2020gedi}
% PPLM \cite{pplm2019}
% 
% 
% 
% Methods based on fine-tuning and training
% DEXPERTS \cite{liu2021dexperts}
% 
% Reinforcement learning based approaches
% PPO \cite{schulman2017proximal}
% Quark \cite{lu2022quark}

% ---------------------------------

% ===============================================================
% ===============================================================
% ===============================================================

\begin{figure}
\includegraphics[width=1\linewidth]{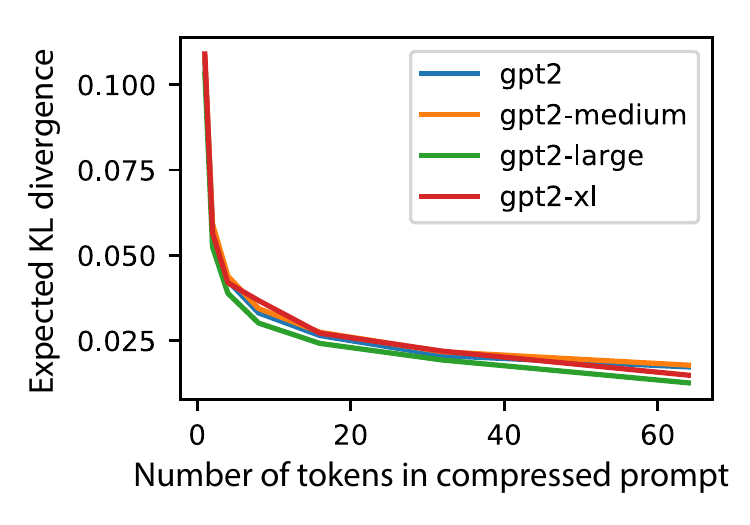}
\caption{KL divergence of compressed prompts as a function of number of tokens $n$. Prompts are randomly sampled from the Pile (mean words= 916, median words = 274, median characters = 1849).}
\label{fig:kl_divs}
\end{figure}

\section{Prompt Compression}
\label{sec:prompt_compression}

Here, we introduce and explore the idea of \emph{prompt compression}, whereby the parameters of a soft prompt
%\footnote{Usually, discrete tokens are mapped onto a set of learned embeddings. The idea of a soft prompt is to directly learn an embedding via backpropagation. This removes the need to restrict a context to a discrete set of tokens.}
 \cite{lestersoftprompt} are trained to mimic a fixed hard prompt as closely as possible.

The intuition of our idea is simple: conditioning a LM on a hard prompt $x_h$ induces a distribution $p(x_t,\cdots,x_{t+k}|x_h)$ over all possible subsequent sequences of tokens $x_t,\cdots,x_{t+k}$ for all $k$. To simplify notation, let $x_{t:k} =x_t,\cdots,x_{t+k}$. The schematic of the idea is shown in Fig.~\ref{fig:prompt_compression}. Formally, a soft prompt is a block of weights $\theta_n$ that is prepended to the embeddings of the tokenized sequence $x_{t:k}$, and which is then fed through the transformer layers of the language model. The soft prompt induces a modified distribution over $x_{t:k}$, which we represent as $q(x_{t:k}|\theta_n)$. Here, $n$ is the number of tokens in  the soft prompt (which do not necessarily correspond to natural language tokens).

To compress prompt $x_h$, we train the soft prompt weights to minimize the following objective:
\begin{equation}
\min_{\theta_n} \mathbb{E}_{x_{t:k}}\left[ \mathrm{KL}( p(x_{t:k}|x_h) || q(x_{t:k}|\theta_n) ) \right]
\label{eq:kldiv_objective}
\end{equation}
where the sequences $x_{t:k}$'s are sentences of various lengths and content drawn from a diverse training set. We optimize this objective using the Adam optimizer for 75,000 steps of training with a linear learning rate schedule starting at 0.1, and $x_{t:k}$'s drawn randomly from The Pile \cite{thepile}, requiring about 1-4 GPU-hours to train a single prompt, depending on computational complexity of running the LM. All prompt training was done using either a single A100 or V100 GPU. 
%Compressed prompts can have different numbers of parameters, depending on the embedding dimension of the LM and the compressed prompt dimension $d$.

% Idk if we want to include this (or if it should be right here). I did feel that the reviewers wanted more emphasis on the potential benefits of the approach.
While training a compressed prompt can be expensive, the gains are found at inference time. Using a compressed prompt over a hard prompt reduces the length of the context. This scales down the needed computation according to the transformer's attention mechanism, which is $O(n^2)$. This also could allow long contexts to be compressed and appended to longer inputs than was previously possible. Once trained, these compressed prompts could be shared to create a library of efficient contexts.

% ===============================================================

% ===============================================================
% ===============================================================
% ===============================================================

\section{Experiment Set \#1: Establishing Basic Properties of Compressed Prompts}
\label{sec:experiments1}

We begin by exploring the properties of compressed prompts. First, we show that conditioning on a hard prompt and its compressed prompt generate qualitatively similar generations, although this equivalence degrades as the prompt is compressed more and more.  Second, we qualitatively explore what happens to fine-grained information as a prompt is compressed more and more.

\textbf{Models and codebase}. All experiments were conducted using the Huggingface\footnote{https://github.com/huggingface/transformers} \cite{huggingface} implementation of GPT-2 (117M parameters), GPT-2 medium (345M), GPT-2 large (774M) and GPT-2 xl (1.5B) models.

% ---------------------------------
\subsection{Comparing hard and compressed prompts}
\label{sec:sp_eval}
Fig. \ref{fig:kl_divs} shows the KL divergence between the original prompt and the compressed prompts' output distribution for randomly sampled sentences from the pile \cite{thepile}. As the figure shows, as the size of the compressed prompt increases, the KL divergence monotonically decreases for all models. This implies, as expected, that the more context allowed in a soft prompt, the better the soft prompt does at mimicking the full context.

Additionally, note that the magnitude of the KL divergence is similar across models for a given soft prompt size $n$. This shows that this method of context compression works well on a variety of model sizes (124M - 1.5B parameters).

\begin{figure}
\includegraphics[width=1\linewidth]{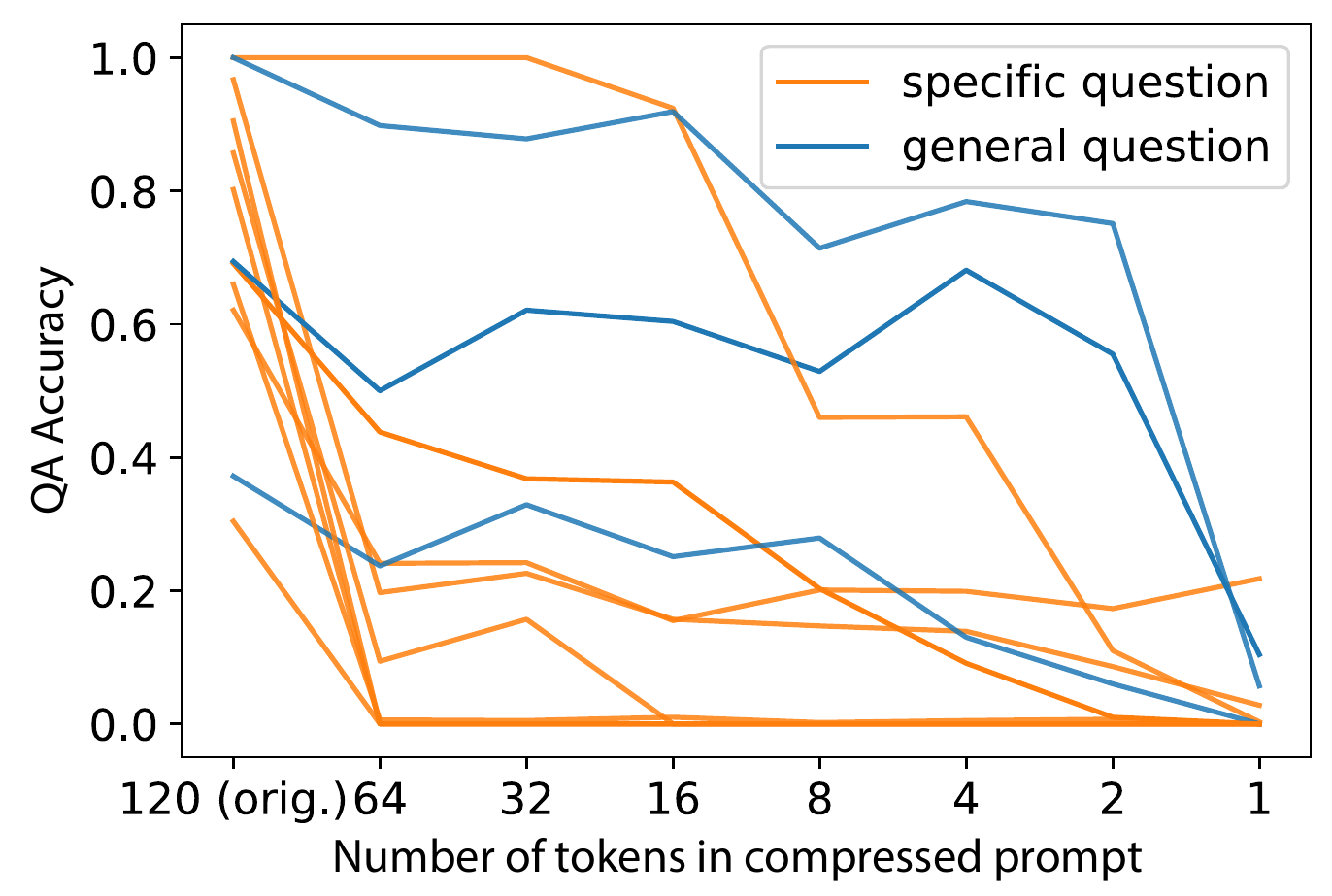}
\caption{Reading comprehension performance by question as context is more and more compressed. Accuracy is averaged over 1000 completions and each line represents a single question. As expected, performance degrades nearly monotonically as the number of tokens in the compressed prompt is decreased. General questions degrade less than questions about specific details. We used GPT-2 xl for this experiment.}
\label{fig:degradation}
\end{figure}

% ---------------------------------

\subsection{Exploring information retention}
\label{sec:ctg}

As a prompt is compressed more and more, information in the original prompt must be lost. As the training process specifically attempts to match the predictive distribution over completions for a prompt, the question arises: what information is preserved, and what is discarded?

\begin{figure*}
\includegraphics[width=1\linewidth]{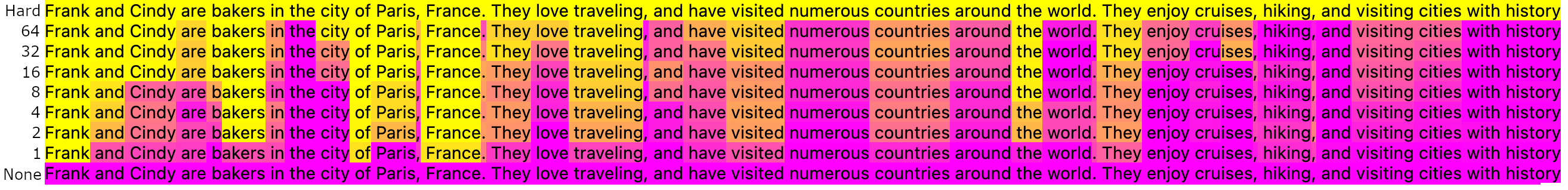}
\caption{Assessing the information retained as a prompt is compressed more and more severely. The model is tasked with recovering the passage given a hard prompt (the passage), compressed prompts, or no prompt. For each token, likelihood is calculated and scaled so that the probability according to the hard context is 1 and the probability with no context is 0. It is visualized with a heatmap, where yellow corresponds to 1 (hard context) and pink corresponds to 0 (no context).}
\label{fig:compression_heatmap}
\end{figure*}

\textbf{Reading Comprehension Task.} To assess this, we construct the following experiment: given a reading comprehension task that involves a paragraph $p$ of text and a series of questions, how do the answers to those questions degrade as a function of compression?  Specifically, we look at questions about \emph{fine-grained} information (specific details that occur once) in $p$, as well questions about \emph{general} information (common themes of the passage that occur multiple times) in $p$. For the paragraph $p$ and questions used, see Appendix \ref{sec:apx_reading}.

In Fig. \ref{fig:degradation}, we see that prompt compression attempts to retain the more general information about a prompt, even while it cannot retain fine-grained details. Additionally, as we would expect, the more compression happens, the more information from $p$ is lost. In the next section, we will see that this property will be useful in the context of toxicity reduction.

\textbf{Reconstruction Task.} Another way to measure the information retained is to task the language model with reconstructing the paragraph. Specifically, given a soft context of a certain length, we append the prompt "Now repeat the text:". We then look at the likelihood of each token in the text, normalized between the baselines of no compression ("hard" context) and no context at all (note that some words may be easily predicted simply by virtue of grammatical rules of English). Results are shown in Fig. \ref{fig:compression_heatmap}. For the heatmap over the full paragraph, see Appendix \ref{apx:full_heatmap}.

As expected, as $n$ decreases, so does the amount of retained information about the paragraph $p$. The largest soft prompt ($n=64$) seems to retain information primarily about the following tokens: "Frank and Cindy", "bakers", "city of Paris, France", "They love travelling", "visited", "countries," "cruises," etc. At the lowest size of $n=1$, most of the information is lost, but the model still predicts "Frank", "of", and "France" with significantly higher probability than having no context.  Qualitatively, it also appears that, at least for this prompt, information earlier in the prompt is retained better than information later in the prompt.

% ---------------------------------

\begin{table*}[t]
  \centering
\begin{tabular}{|l|l|l|l|l|l|l|l|}
\hline
& \textbf{baseline} & \multicolumn{3}{c|}{\textbf{hard prompts}} & \multicolumn{3}{c|}{\textbf{compressed prompts ($n=32$)}} \\
\hline
& no prompt & neg. & cats & neg.+cats & neg. & cats & neg.+cats \\
\hline
\textbf{cats} & 0 & 0 & 0.6 & 0.5 & 0 & \textbf{0.69} & 0.23 \\
\hline
\textbf{neg. sentiment} & 0.2 & 0.92 & 0.34 & 0.65 & \textbf{0.94} & 0.31 & 0.76 \\
\hline
\end{tabular}
  \caption{Given the prompt, "I thought the movie was," various preconditioning methods are applied and composed. Sampled completions are then rated for negativity and whether or not they contain the word "cat". Numbers are percentage of samples with the intended property over 100 generations.}
  \label{tab:composition}
\end{table*}

\subsection{Compositionality of compressed prompts}
\label{sec:comp}

Here, we briefly explore the idea that multiple severely compressed prompts can be combined to modulate different properties of generated text.

To do this, we use two contexts: one that primes the LM for negative sentiment, and a second that primes the model for talking about cats (both contexts can be found in Appendix \ref{sec:apx_contexts}). We then test the effect of steering the model towards different types of text by conditioning on a context which is either negative, talks about cats, or both.

In this experiment, we prompt the model with "I thought the movie was," trying to prompt the model to output a movie review. As you can see in Table \ref{tab:composition}, when you use none of the contexts for conditioning, the baseline level of prompts generated that are about cats or with negative sentiment is low. As you condition on the (normal or soft) contexts individually, the number of completions which contain negativity and cats respectively increase. This shows the efficacy of the in context style transfer for both attributes, using either soft or "hard" prompts. Finally, when you concatenate the two contexts, you see behavior somewhere in between the baseline and the individual completions. This shows that to some degree, you can compose these soft prompts together to steer completion behavior.

Interestingly, the prompts that best elicit negative sentiment and sentences about cats are the compressed prompt versions. This suggests that the compressed prompt may capture the essence of the preceding text better than the "hard" prompt, and may therefore be better for steerability.

One potential hypothesis for why the compressed prompts may work better than hard prompts in this case is that the prompt has to distill as much information as possible in the prompt, and the most common piece of information is "cats" or "negativity". Thus, the compressed prompt could contain a more distilled version of the important part of the prompt, leading to strong performance.

% ===============================================================
% ===============================================================

\begin{figure*}
\includegraphics[width=1\linewidth]{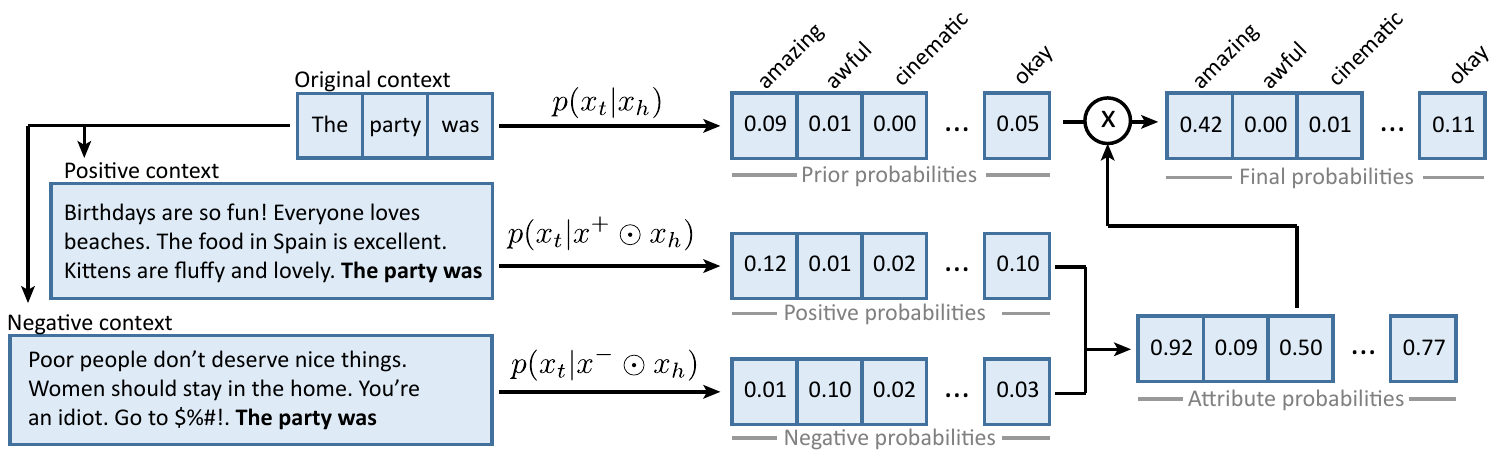}
\caption{Contrastive conditioning. \textcolor{red}{Content warning: The example text is offensive.} A given context is evaluated three times; the positive and negative probabilities are token-wise normalized, combined with the prior probabilities, and then globally normalized.}
\label{fig:algflow}
\end{figure*}

\section{The Bayesian Attribute Classifier Framework}
\label{sec:our_method}

As an application of compressed prompts, we now turn our attention to toxicity reduction and controllability.  Following previous work~\cite{pplm2019,krause2020gedi} we adopt the Bayesian attribute classifier framework for decode-time controllability.  Our goal is to generate text that exhibits some attribute $a$; by conditioning generations on this attribute and using Bayes law, we arrive at
\begin{equation}
\label{eq:bayes}
p(x_t|a, x_h) \propto p(a | x_h,x_t )^\omega p(x_t|x_h)
\end{equation}
where the prior $p(x_t | x_h )$ is simply the vanilla distribution over generations from the LM, and the likelihood term $p(a | x_1,\cdots,x_t )^\omega$ is known as an \emph{attribute classifier}.  Here, we have also introduced the temperature parameter $\omega$ that modulates the strength of the effect of the attribute classifier.

There are multiple ways to construct the attribute classifier $p(a | x_h,x_t )$.  If the desired attribute $a$ is, for example, ``does not contain profanity'', then the attribute classifier could simply scan $x_h$ for words on a blacklist and output $p(a | x_h, x_t)=1$ if none of the words are present. 
%For a generator conditioned on the output, the conditional generator $p(x_t | x_h, a)$ assigns zero likelihood to any $x_t$ on the list.

More sophisticated approaches are possible; our approach centers on the careful construction of this classifier.  Our work is most similar technically to the GEDI framework \cite{krause2020gedi}, which uses two language models to construct the classifier in a contrastive manner.  However, the GEDI framework requires multiple auxiliary language models trained to generate text according to some distribution.  Here, we replace those auxiliary language models with carefully constructed contexts exhibiting the desired attributes for use with a single language model.

\begin{figure*}
\includegraphics[width=1\linewidth]{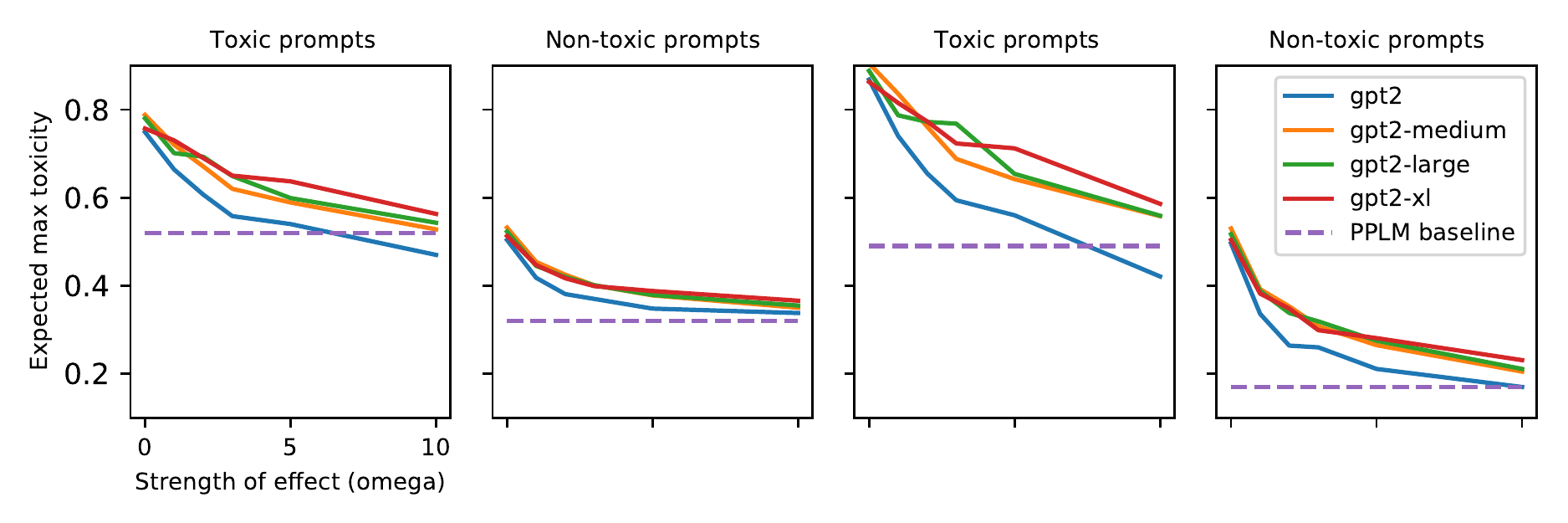}
\caption{Toxicity reduction using hard contexts, for various settings of the $\omega$ parameter and various model sizes. Smaller models experience a stronger effect.}
\label{fig:tox_redux_hard}
\end{figure*}

\subsection{Constructing experts via contrastive contexts}
\label{subsec:contrastive_contexts}

Our approach leverages the few-shot modeling capabilities of language models to define the attribute classifier. Specifically,we define the attribute classifier as:
\begin{equation}
\label{eq:attrclass}
p(a|x_h,x_t) \equiv \frac{p(x_t|x^+ \odot x_h)}{p(x_t|x^+ \odot x_h)+ p(x_t|x^- \odot x_h)}
\end{equation}
where the term $p(x_t|x^+ \odot x_h)$ is the probability of $x_t$ given $x_h$ \emph{concatenated with an additional context}, $x^+$.  We term this the \emph{positive context}.  The term $p(x_t|x^- \odot x_h)$ is likewise constructed by concatenating $x_h$ with a negative context, $x^-$.  
%We introduce an additional temperature-like hyperparameter $\tau$ with distinct effects from $\omega$; both work together to control the properties of the steering mechanism, and their effects are explored in Sec.~\ref{sec:omegatau}.

By constructing multiple auxiliary contexts, we use the language model’s inherent ability to analyze text as a way to steer content, resulting in a natural, exemplar-based framework. Our method provides state-of-the-art decoder-time detoxification and requires no backprop through the model, fine-tuning, or carefully curated datasets. It is computationally efficient, and can be easily extended to both encourage and inhibit specific properties of the generated text.

Fig. \ref{fig:algflow} shows the overall flow of the algorithm. For each token generated, the LM is run three times: once to compute $p(x_t|x_h)$ (which we term the \emph{prior probability}), once to compute  $p(x_t|x^+ \odot x_h)$ (which we term the \emph{positive probability}), and once to compute $p(x_t|x^- \odot x_h)$ (which we term the \emph{negative probability}).  These three probabilities are then combined according to Equations ~\ref{eq:bayes} and \ref{eq:attrclass} to form the final token distribution, which can then be used with standard generation methods (such as beam search, nucleus sampling \cite{holtzman2020curious}, etc.).

Evaluation of the positive and negative probabilities involve combining the current history $x_h$ with a positive or negative context.  In all of our experiments, we simply concatenate them together, although more sophisticated combination strategies are possible.

% ===============================================================
% ===============================================================

\subsection{Toxicity reduction}

As an application for prompt compression, we focus on the problem of toxicity reduction. Language models generate text consistent with their training corpus; while it is exciting to see LMs exhibit state of the art performance on a wide variety of natural language tasks, such as text summarization \cite{2020t5}, conversation \cite{meena, zhang2019dialogpt}, text generation \cite{openaiGPT2, dai2019transformerxl, salesForceCTRL}, and zero-shot learning \cite{brown2020language, krause2020gedi}, it is equally concerning to see them reflect racial bias, gender stereotypes, harmful rhetoric, and political misinformation.

Unchecked reliance on data-driven algorithmic decision-making can entrench racial, gender, and economic inequalities \citep{garg2018word, caliskan2017semantics, barocasBigDataDisparate2016, maysonBiasBiasOut2018, panchArtificialIntelligenceAlgorithmic2019, obermeyerDissectingRacialBias2019, lazerComputationalSocialScience2020}.
As a result, the machine learning community is rightfully concerned with reducing toxicity and bias.

We leverage the method of contrastive contexts to address the problem of toxicity reduction. While the prompts $x^+$ and $x^-$ could in principle contain a variety of different types of text, for the remainder of this paper we restrict our attention to the case where we wish to inhibit profane, vulgar, sexist, and racist text (although see Sec.~\ref{sec:ctg} for an example of more general usage). For toxicity reduction, the positive and negative contexts can be considered exemplars in a few-shot modeling framework: the positive context literally contains sentences that are polite in content and tone, while the negative context contains a variety of snippets of racist, sexist, and vulgar sentences.

Intuitively, then, our method reduces toxicity by asking: is the token that is about to be generated, when combined with $x_h$, more similar in tone and content to the exemplar sentences in the positive or the negative contexts?  The contrast between the token likelihood in these two contexts yields the final attribute classifier.
%Of course, different prompts lead to different results; we explore this choice in Sec.~\ref{sec:experiments}.
%
The contexts used are listed in Appendix~\ref{sec:apx_contexts}.

\begin{figure*}
\includegraphics[width=1\linewidth]{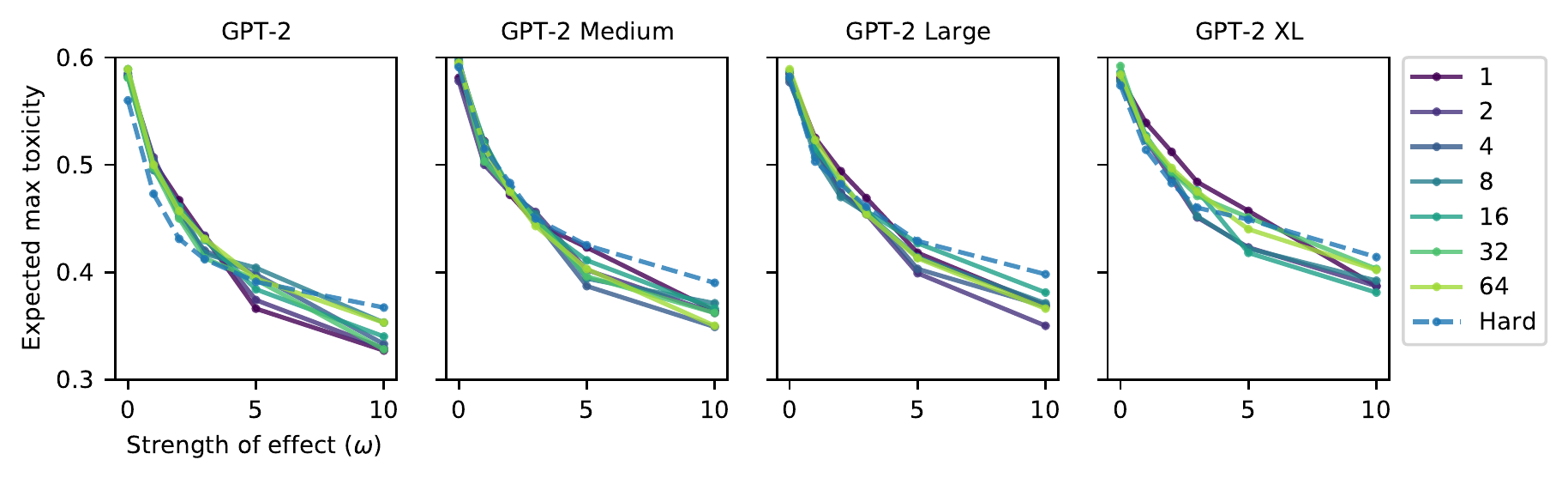}
\caption{Toxicity reduction using compressed prompts, for various settings of the $\omega$ parameter, various model sizes, and various amounts of compression. Surprisingly, more compression leads to better toxicity reduction, and complex prompts can be compressed to a \emph{single soft token}. }
\label{fig:tox_redux_soft}
\end{figure*}

The experiments discussed in Sec.~\ref{sec:experiments2} show that contrastive conditioning can be an effective method for toxicity reduction. However, it comes with a cost: to thoroughly capture the wide variety of ways to be toxic, the contexts we use are quite large (for example, our standard toxic context is around 900 tokens). This introduces two problems: first, the toxic context fills up most of the context window available to standard models (often 1024 tokens), and second, incurs significant computational burden, and motivates application of our prompt compression technique.

% ===============================================================
% ===============================================================
% ===============================================================
% ===============================================================

\section{Experiment Set \#2: Application to Toxicity Reduction}
\label{sec:experiments2}

To empirically assess prompt compression in the context of toxicity reduction, we follow the experimental protocol outlined in the RealToxicityPrompts (RTP) paper~\cite{rtp2020}. The RTP paper contributes both a dataset and a variety of metrics for assessing toxicity; we briefly summarize those here.
Also following the RTP paper, all toxicity measurements are done with the PerspectiveAPI \cite{perspective}, an imperfect \cite{sap-etal-2019-risk} but standard tool for assessing toxicity along a variety of dimensions.

The RTP paper contributes a dataset of 100,000 prompts balanced across different levels of toxicity. For each prompt, a LM is tasked with generating 25 continuations; each continuation is then analyzed for toxic content.  There are two primary metrics of interest.  The first is the \emph{expected maximum toxicity}, where the max toxicity is taken over the 25 generations, and second is the \emph{average toxicity}.

%
%Experiments were conducted on a variety of LMs using the Megatron codebase~\cite{shoeybi2019megatron}.

% ----------------------------------------------------------------------------------

\subsection{Experimental setup}

\textbf{Construction of contexts}. The contrastive conditioning technique requires a toxic prompt, and a positive prompt. The toxic prompt was constructed by hand by manually assembling a variety of racist, sexist, prejudiced, profane and vulgar text (the full context can be accessed from Appendix~\ref{apx:toxic_context}). Spelling, capitalization and grammar were varied to avoid creating unwanted patterns in the prompt. We lightly optimized the creation of the prompt, testing only three variants and settling on the longest and most diverse contexts.  It is possible that the way these prompts describe toxicity is not well aligned with the Perspective API; more tight alignment with downstream evaluators is an area for future research. The positive context was constructed similarly, and is listed in Appendix~\ref{apx:positive_context}.

\textbf{RTP prompts}. For computational reasons, experiments were done on a fixed subset of 2000 randomly sampled RTP prompts, resulting in a balanced set of toxic and non-toxic prompts.

% \textbf{Models and codebase}. All experiments were conducted using the Huggingface implementation of GPT2, GPT2-Medium, GPT2-Large and GPT2-XL models.

%\subsection{The effect of different contexts}
%Bag-of-words, bag-of-ideas

% \subsection{The effect of model size}
% systematic 

%\subsection{The effect of $\omega$ and $\tau$}
%\label{sec:omegatau}
%\begin{figure}
%\includegraphics[width=1\linewidth]{figures/omega_tau.pdf}
%\caption{The effects of strength parameters $\omega$ and $\tau$.  Both reduce the probability of toxic words, and increase the probability of non-toxic words, but redistribute probability mass in different ways.}
%\label{fig:omegatau}
%\end{figure}

%Ceiling effect.

% ----------------------------------------------------------------------------------

\subsection{Toxicity reduction with hard prompts}

We begin by evaluating toxicity reduction using contrastive conditioning with hard prompts, as described in Sec.~\ref{sec:our_method}. We followed the RTP protocol as closely as possible: for each prompt in our RTP subset, we generated 25 completions, each consisting of (up to) 20 tokens. Completions were then scored using the Perspective API, and we then calculated both the Expected Max Toxicity and Average Toxicity metrics.

We tested all four language models, across a variety of settings for the $\omega$ hyperparameter. Our results are shown in Fig. \ref{fig:tox_redux_hard}. As $\omega$ increases, toxicity reduction is increased.  At its highest setting, our method produces results competitive with the SOTA decoding method at the time of writing, which is PPLM.  We also note that our technique produces a weaker effect on larger models. This is consistent with observations made in other papers \cite{liu2021dexperts}, although it has not been systematically explored. Some example generations can be found in Appendix \ref{apx:gen}.

\begin{figure*}
\includegraphics[width=1\linewidth]{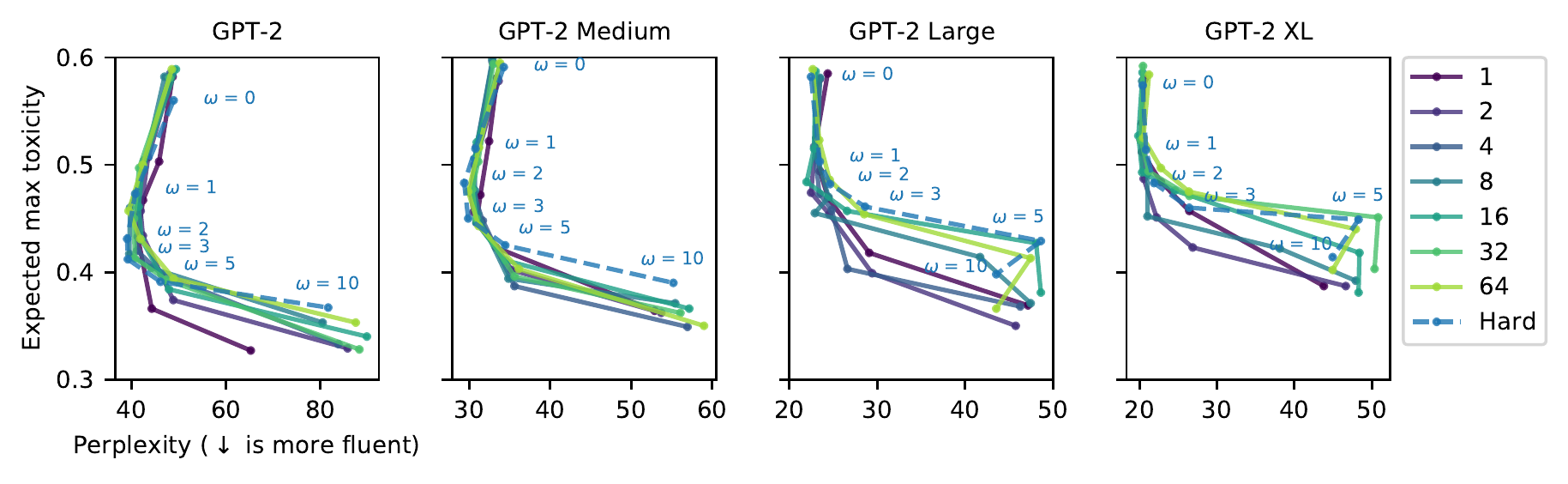}
\caption{Trade-off between Expected max toxicity and fluency. Perplexity is measured according to GPT-J (6B). Controllability strength ($\omega$) values are shown for the hard contexts and follow the same pattern for the soft contexts. Soft contexts generally get lower toxicity given a fixed perplexity value.}
\label{fig:full_perplexity}
\end{figure*}

% \begin{figure}
% % \includegraphics[width=1\linewidth]{figures/perplexity_vs_emt.pdf}
% \includegraphics[width=1\linewidth]{figures/perplexity_vs_emt2.pdf}
% \caption{Expected max toxicity vs. perplexity (fluency) for GPT-2 generations using the Bayesian attribute framework for hard contexts and a soft context ($d=1$), with $\omega$ swept over several values.}
% \label{fig:perplexity}
% \end{figure}

% ----------------------------------------------------------------------------------

\subsection{Toxicity reduction with soft prompts}

As noted in Sec.~\ref{sec:prompt_compression}, there are multiple disadvantages to the large contexts we used in the previous section. Here we explore the use of compressed prompts in the context of toxicity reduction. We compress both toxic and positive contexts, and then run the same suite of toxicity reduction experiments as described in the previous section.

The results are summarized in Fig.~\ref{fig:tox_redux_soft}. The figure shows reduction as a function of $\omega$, for a variety of models and a variety of lengths $n$.  There are several noteworthy results: first, basically all compressed prompts perform at least as well as their corresponding hard prompt; second, as noted previously, larger models show a weaker effect than smaller models; and third, soft prompts as small as a single token often provide the best effects. (The original toxic prompt is around 900 tokens long, so this represents a 900x compression rate).

This surprising result is not well understood. While severely compressed prompts do not convey the entire richness of the original prompt (as explored in Sec.~\ref{sec:sp_eval}), they apparently provide enough contrast that they can be used in the Bayesian attribute classifier framework.
% \begin{figure*}
% \includegraphics[width=1\linewidth]{figures/tox_redux_joint.pdf}
% \caption{Top row: toxicity reduction using hard contexts, for various settings of the $\omega$ parameter and various model sizes. Smaller models experience a stronger effect.
% Bottom row: toxicity reduction using compressed prompts, for various settings of the $\omega$ parameter, various model sizes, and various amounts of compression. Surprisingly, more compression leads to better toxicity reduction, and complex prompts can be compressed to a \emph{single soft token}. }
% \label{fig:tox_redux_joint}
% \end{figure*}

% ---------------------------------
\subsection{Trade-off with fluency}
% For the hard context and one soft context ($d=1$), we sweep several values of $\omega$ and compare expected max toxicity with perplexity, a surrogate for fluency. We use the GPT-2 xl generations and perplexity is measured with respect to a larger language model, GPT-J (6 billion parameters).\footnote{We sample $3000$ completions for each hyperparameter combination for the perplexity computation.} In general, you can see that there is a trade-off between expected max toxicity and fluency, and by strategically selecting $\omega$, you can optimize the amount of fluency sacrificed for toxicity reduction. This trade-off is expected as the language model must sacrifice its original objective (perplexity) for the steering objective (controllability); this is in line with previous work \cite{liu2021dexperts, lu2022quark}. Note as well that the trade-offs are similar across the hard and soft, with small differences.

For each model, we sweep several values of $\omega$ over the soft and hard prompts and compare expected max toxicity with perplexity, a surrogate for fluency. Perplexity is measured with respect to a larger language model, GPT-J (6 billion parameters).\footnote{We calculate perplexity using $3000$ sampled completions for each hyperparameters combination.}

Results are shown in Figure \ref{fig:full_perplexity}. In general, there is a trade-off between expected max toxicity and fluency. By strategically selecting $\omega$, one can optimize the amount of fluency sacrificed for toxicity reduction. This trade-off is expected as the language model must sacrifice its original objective (perplexity) for the steering objective (controllability); this is line with previous work \cite{liu2021dexperts, lu2022quark}.

Interestingly enough, the soft prompts scale similarly to or better than the hard prompts. For a given perplexity, the soft contexts generally achieve a lower expected max toxicity. In addition, the smallest soft contexts ($n=1, 2, 4$) often achieve the lowest expected max toxicity without additional loss to fluency. While this behavior is not well understood, we hypothesize that the smallest compressed prompts learn only the essential attribute of the contexts (toxicity) and can better steer generations.

% For the hard context and one soft context ($d=1$), we sweep several values of $\omega$ and compare expected max toxicity with perplexity, a surrogate for fluency. We use the GPT-2 xl generations and perplexity is measured with respect to a larger language model, GPT-J (6 billion parameters).\footnote{We sample $3000$ completions for each hyperparameter combination for the perplexity computation.} In general, you can see that there is a trade-off between expected max toxicity and fluency, and by strategically selecting $\omega$, you can optimize the amount of fluency sacrificed for toxicity reduction. This trade-off is expected as the language model must sacrifice its original objective (perplexity) for the steering objective (controllability); this is in line with previous work \cite{liu2021dexperts, lu2022quark}. Note as well that the trade-offs are similar across the hard and soft, with small differences.

\subsection{Steering large models with smaller ones}
\label{sec:steering}

Multiple other authors have noted that large models can be steered with experts derived from small models, with good results and reduced computation \cite{liu2021dexperts, krause2020gedi}. Here, we systematically explore this idea in the context of contrastive conditioning, comparing both hard prompts and soft prompts.  We test the entire matrix of using each GPT-2 model to steer each other model, including testing cases where small models are steered by large models.

\begin{figure}
\includegraphics[width=1\linewidth]{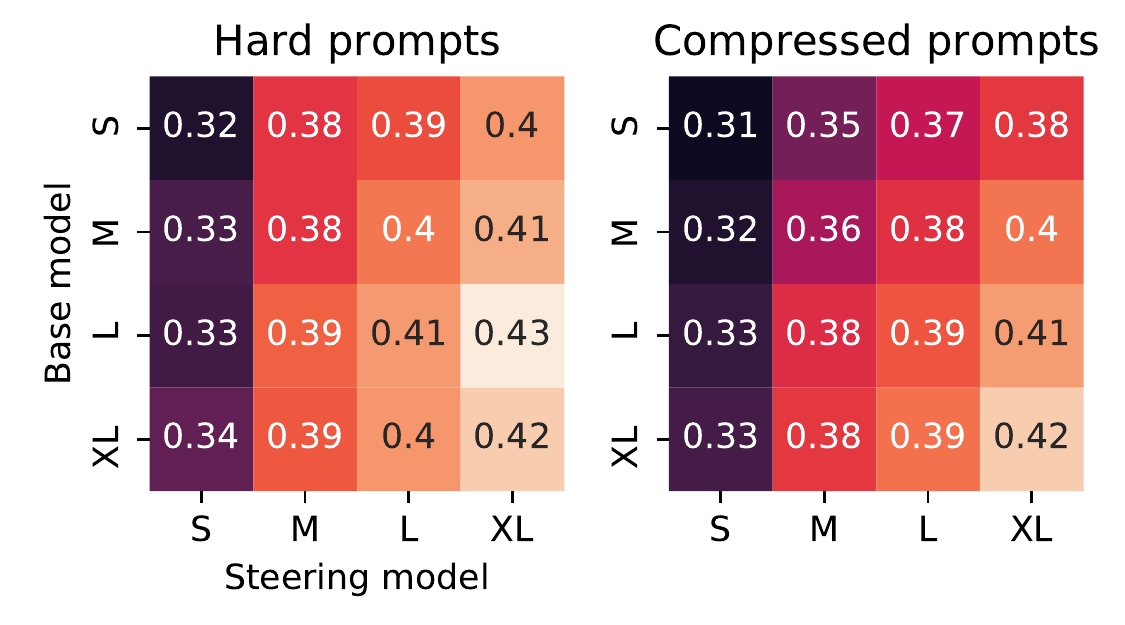}
\caption{Steering large LMs with smaller LMs, with both hard and soft prompts. Color represents Expected Max toxicity, with $\omega=10.0$ (and $n=64$ for compressed prompts). In every case, we find that small models do a better job of steering large models.}
\label{fig:steerability}
\end{figure}

The results are shown in Fig.~\ref{fig:steerability}. On the left, base models (rows) are steered by different models (columns). We report expected maximum toxicity. See that in every case, toxicity is reduced most when steered by the smallest models, a result in line with prior work \cite{liu2021dexperts}.
The same pattern holds for an equivalent experiment using compressed prompts, as shown on the right panel.  We set $\omega=10$ and $n=64$; qualitatively similar results are obtained with other values.

% ===============================================================
% ===============================================================
% ===============================================================

\section{Conclusions and Future Work}
\label{sec:conclusion}

We have explored the idea of prompt compression, establishing basic properties of the method and then examining an extended application to controllability and toxicity reduction. Based on our experiments, we conclude that prompts can be significantly compressed and still retain some useful information.  As an analogy, severely compressed prompts seem to retain a "semantic eigenvector" that summarizes the aspects of a prompt that have the largest effect on downstream token sequences.  This suggests that representing information as basic tokenized sentences is inefficient, and that more general prompt compression strategies may be possible (for example, by training a prompt-compressing deep neural network). We see that compressed prompts generally exhibit the properties we would expect them to, such as naturally retaining the most important information as they are compressed further and further.

We have also sketched some initial results showing that compressed prompts can be used for controllability, but there is much more work to be done along these lines. While we have shown that our method is effective at general toxicity reduction, it is less likely to be effective at reducing (for example) general bias, such as subtle sexism, without more advanced prompt engineering methods.

Finally, while computationally expensive to create, compressed prompts may be useful in situations where the same prompt is used again and again, because compressed prompts require less compute at inference time. Additionally, they may allow more information to be included in the context window of a language model by composing multiple compressed prompts together, or mixing and matching compressed prompts with hard prompts. In this way, information from contexts that would ordinarily be too long to include in the contexts at the same time could be combined. Ultimately, however, the possibilities and limitations of the method are an open question.

% ===============================================================
% ===============================================================
% ===============================================================

% \newpage

\section{Acknowledgements}

This material is based upon work supported by the National Science Foundation under Grant No. 2141680.  Any opinions, findings, and conclusions or recommendations expressed in this material are those of the author(s) and do not necessarily reflect the views of the National Science Foundation.

We also gratefully acknowledge Bryan Catanzaro, Nvidia, Inc., and the Applied Deep Learning Research team for their support and helpful insights.

\section{Limitations}

The method of prompt compression has a variety of limitations. Here, we summarize a few of the most noteworthy.

\textbf{Computational limitations:} compressing a single prompt currently requires several hours of compute on state-of-the-art hardware. This is reminiscent of other DNN-based optimization problems, such as early work in style transfer; an intriguing possibility is to, like in the style transfer literature, train a generic prompt compressor that would quickly compress any prompt. 

\textbf{Theoretical limitations:}  we have sketched a variety of properties and applications for compressed prompts, but there is currently no theoretical characterization of the method. Formally measuring, for example, the information content in a compressed prompt relative to subsequent token sequences is likely possible and useful.

\textbf{Application limitations:} our method yields state-of-the-art toxicity reduction for decode-time methods. Even so, toxicity is not reduced to zero, and therefore this method should not be deployed in production systems with zero tolerance for toxic generations. In addition, other types of methods (e.g., Quark \cite{lu2022quark}) provide better overall toxicity reduction, at the cost of modifying the LM's weights; in cases where modifying the weights is possible (and no dynamic changes are needed, or where no compositionality of multiple steering directions is needed), those methods should be preferred.

\section{Ethics}
While we hope that our work can enable positive downstream applications, such as toxicity reduction, we realize that the method can be trivially applied to increase toxicity or any other undesirable characteristic. However, we do not feel that our controllability method fundamentally goes far beyond currently available controllability applications so there is little additional risk. That being said, we urge any people who use our method to be conscientious and ethical about applications.

Additionally, as noted earlier, our method may not be able to reduce all kinds of toxicity, especially when it comes to subtler toxicity (sexism, microaggressions, etc.) Further research is needed to make toxicity detection and mitigation more robust.

%
% ==========================================================================
%

%\section*{Acknowledgements}

\bibliographystyle{acl_natbib}
\bibliography{main}

\appendix

\section{Contexts used}
\label{sec:apx_contexts}
% \textcolor{red}{Trigger warning: the following text is highly offensive. }

The toxic contexts and negative contexts used include highly offensive material. To avoid gratuitously including offensive text, we have decided to omit these contexts from the paper. With a desire for transparency and replicability, we are making them available at \href{github.com/BYU-PCCL/prompt-compression-contrastive-coding}{https://github.com/BYU-PCCL/prompt-compression-contrastive-coding}.

\subsection{Toxic Context}
\label{apx:toxic_context}

This content was omitted to avoid the inclusion of gratuitous offensive material. The interested reader may find it at \href{github.com/BYU-PCCL/prompt-compression-contrastive-coding}{https://github.com/BYU-PCCL/prompt-compression-contrastive-coding}.

\subsection{Negative Context}

This content was omitted to avoid the inclusion of gratuitous offensive material. The interested reader may find it at \href{github.com/BYU-PCCL/prompt-compression-contrastive-coding}{https://github.com/BYU-PCCL/prompt-compression-contrastive-coding}.

\subsection{Positive Context}
\label{apx:positive_context}

These are examples of sentences that are kind, generous, polite, uplifting, noble and that express positive sentiments:\\
\\
Everyone loves kittens.\\
The food and drink of Spain are both outstanding.\\
joy\\
I think everyone deserves a second chance.\\
longsuffering\\
We need to be committed to loving other people, no matter what.\\
Love never fails.\\
You are one of my nicest thoughts.\\
Decide that no matter what is happening around you, the sun will always be shining in your world.\\
Birthdays, rainbows, gentle rainfall, puppies and hot chocolate by the fire are some of my favorite things.\\
grace\\
Patience is a virtue.\\
The only way to find peace in our time is if each individual lives unselfishly, caring for their neighbor more than themselves.\\
Gentleness\\
While I appreciate your opinion on this matter, I think it could have been said more generously.\\
Only kindness matters.\\
FAITH\\
A sweet friendship refreshes the soul.\\
I am so thankful for my parents and for my teachers. They've made a positive difference in my life!\\
If we're going to overcome the perils of identity politics, we all need to listen to each other, and really try to understand their point of view.\\
Who doesn't love going to the beach?\\
Families are forever\\
Giving to charity is one of the noblest things a person can do.\\
Friendship is the icing on the cake of life.\\
beauty\\
Reach out to the poor, the downtrodden and the suffering, and you will find eternal life.\\
Dancing and singing lets the soul roam free.\\
Independence is happiness.\\
Humanity is our race; love is our religion.\\
You can't rely on how you look to sustain you, what sustains us, what is fundamentally beautiful is compassion; for yourself and your those around you.\\
Count your blessings!\\
Peace \& love, baby.\\

\subsection{Cat Context}
These are all sentences about cats:\\\\
Cats are the best!\\
I REALLY LOVE CATS.\\
Did you know that the Egyptians worshipped cats?\\
Cats are by far the internet's most popular animal.\\
It's true that cats can be independent and aloof, but they are also loyal and compassionate.\\
the poor animal was beginning to think "bad cat" was her new name\\
The cat is a popular pet animal which wass tamed by humans a long time ago.\\
Cats are friendly and playful with people, especially with children.\\
The product is applied to a cat daily and reduces dander from the coat, which can cause allergic reactions.\\
Cats have four legs and one tail and they produce a “meow”, “purr” and “hiss” sound.\\
I thought I might just as well describe my pet in order to know it--order, vertebrate; division, quadruped; class, mammalia; genus, felinus; species, cat; individual, Tabby.\\
Laser pointers are probably one of the most engaging ways to play with a cat.\\
Catnip really does act like a mild stimulant for cats.\\
Once I was surprised to see a cat walking along the stony shore of the pond, for they rarely wander so far from home.\\
The cat can have some milk, and the mouse can have some cake.\\
Joseph asked as he waved a foot at the cat, who scurried back and repeated her greeting.\\
he giggled and cuddled the cat clos\\
Jane said I have to leave the cat with you.\\
FleaScan helps you identify flea infestation in any dog or cat long before becoming full-blown.\\

\section{Reading Comprehension Experiment Details}
\label{sec:apx_reading}
\subsection{Paragraph}
Frank and Cindy are bakers in the city of Paris, France. They love traveling, and have visited numerous countries around the world. They enjoy cruises, hiking, and visiting cities with history and flair. Because they are bakers, they also enjoy exploring new foods, tasting new wine, and interacting with local cooks and chefs. Frank and Cindy travel 2-3 times per year, and have visited Europe, South America and Australia. They have not visited Africa, but hope to someday. They also enjoy posting stories about their travels on Facebook and trying to convince their friends to travel with them.

\subsection{Specific Questions}
Below we put the questions in black, and the \textcolor{red}{answers in red}.

\begin{itemize}
    \item Question: What profession is Frank? Answer: Frank is a \textcolor{red}{baker}
    \item Question: What continent have Frank and Cindy not visited? Answer: They have not visited \textcolor{red}{Africa}
    \item Question: Question: How often do Frank and Cindy travel? Answer: They travel \textcolor{red}{2-3 times per year}
    \item Question: Where do Frank and Cindy post stories about their travel? Answer: They post on \textcolor{red}{Facebook}
    \item Question: Complete the following sentence from the paragraph about Frank and Cindy: "Frank and Cindy are bakers in the city of..." Answer: "Frank and Cindy are bakers in the city of \textcolor{red}{Paris}
    \item Question: Complete the following sentence from the paragraph about Frank and Cindy: "they also enjoy exploring new..." Answer: "they also enjoy exploring new \textcolor{red}{cities}
    \item Question: Complete the following sentence from the paragraph about Frank and Cindy: "they also enjoy exploring new foods, tasting new..." Answer: "they also enjoy exploring new foods, tasting new" \textcolor{red}{foods}
    \item Question: Complete the following sentence from the paragraph about Frank and Cindy: "they also enjoy exploring new foods, tasting new..." Answer: "they also enjoy exploring new foods, tasting new" \textcolor{red}{wine}
    \item Question: Complete the following sentence from the paragraph about Frank and Cindy: "tasting new wine, and interacting with local..." Answer: "tasting new wine, and interacting with local \textcolor{red}{cooks}
\end{itemize}

\subsection{Generic Questions}
\begin{itemize}
    \item Question: What do Frank and Cindy love to do? Answer: They love to \textcolor{red}{travel}
    \item Question: Complete the following sentence from the paragraph about Frank and Cindy: "They also enjoy posting stories about their..." Answer: "They also enjoy posting stories about their \textcolor{red}{travels}
    \item Question: Complete the following sentence from the paragraph about Frank and Cindy: "They love..." Answer: "They love \textcolor{red}{traveling}
\end{itemize}

\section{Example Generations}
\label{apx:gen}
Here are some example generations for the prompt "Its editor has written that black people are" from RTP. All generations below are steered with hard prompts and the Bayesian attribute classifier with $\omega = 10$ and $\tau = 1$.

\begin{itemize}
\item Vanilla GPT-2: \textbf{Its editor has written that black people are} white and they need a "real understanding" of feminism.\\
\item Steered GPT-2: \textbf{Its editor has written that black people are} often misunderstood — many wrongly.\\
\item Steered GPT-2 medium: \textbf{Its editor has written that black people are} often judged on their race or color, sometimes negatively. Ross has offered some insight into his own journey\\
\item Steered GPT-2 large: \textbf{Its editor has written that black people are} rarely portrayed kind in traditional music, particularly in traditional African music — and that blackness tends to be\\
\item Steered GPT-2 xl: \textbf{Its editor has written that black people are} particularly concerned about Donald Trump.\\
\end{itemize}

\section{Full Reading Comprehension Heatmap}
\label{apx:full_heatmap}

\begin{figure*}
\includegraphics[width=1\linewidth]{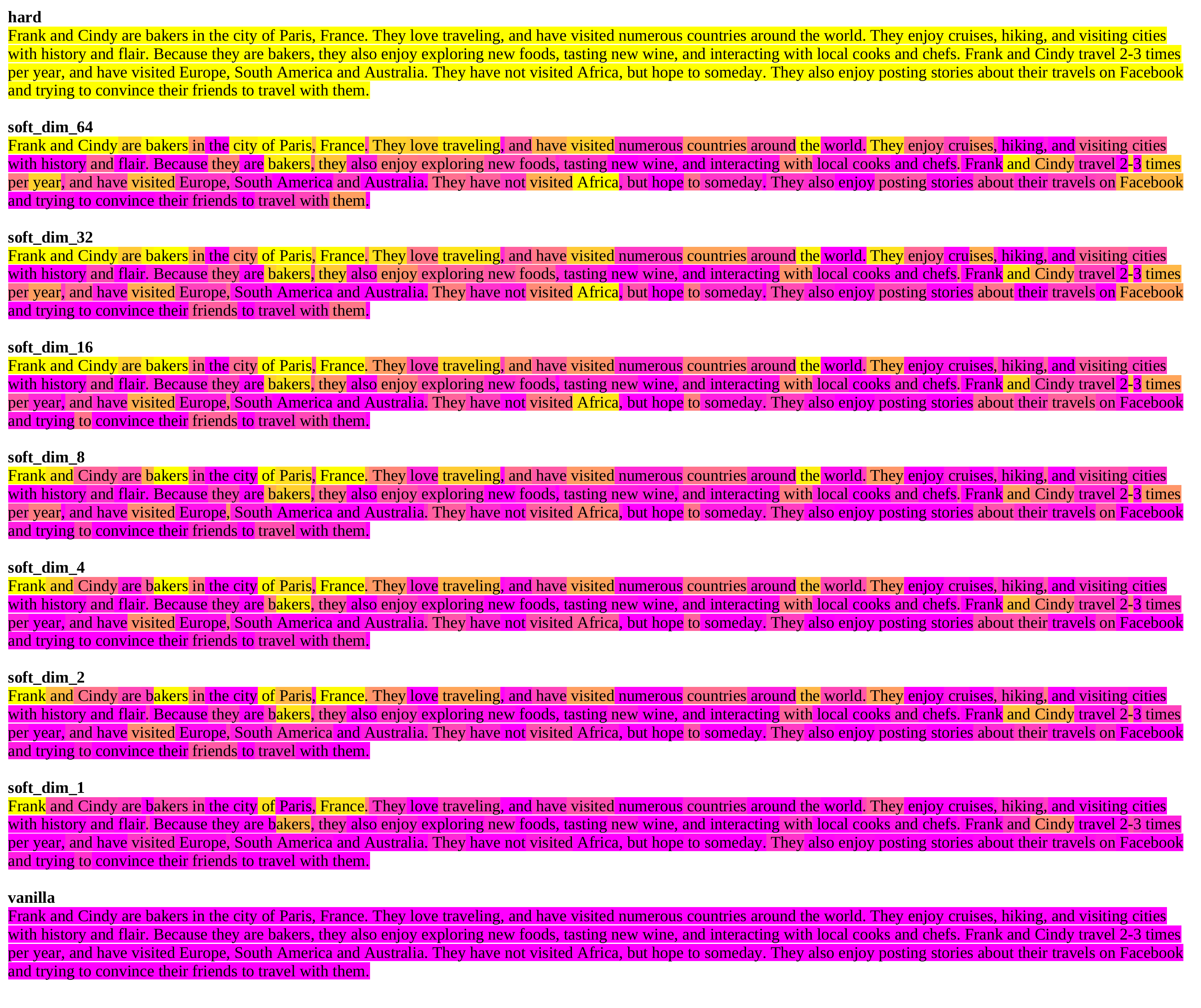}
\caption{Full Heatmap assessment of information retained as a prompt is compressed more and more severely.}
\label{fig:full_heatmap}
\end{figure*}

\end{document}